\documentclass{article}

\usepackage{PRIMEarxiv}

\usepackage[utf8]{inputenc} 
\usepackage[T1]{fontenc}    
\usepackage{hyperref}       
\usepackage{url}            
\usepackage{booktabs}       
\usepackage{amsfonts}       
\usepackage{nicefrac}       
\usepackage{microtype}      
\usepackage{lipsum}
\usepackage{fancyhdr}       
\usepackage{graphicx}       
\graphicspath{{media/}}     
\usepackage{amsmath}
\usepackage{amssymb}
\usepackage{mathtools}
\usepackage{amsthm}
\usepackage{adjustbox}
\usepackage{caption}
\usepackage{subcaption}
\usepackage{algorithm}
\usepackage{algpseudocode}

\title{Topology-Aware Multiscale Mixture of Experts for Efficient Molecular Property Prediction}

\author{
  Long D. Nguyen \\
  School of Mathematics and Statistics \\
  Victoria University of Wellington \\
  Wellington, New Zealand\\
  \texttt{duylong.nguyen@vuw.ac.nz} \\
  \And
  Kelin Xia \\
  School of Physical and Mathematical Sciences \\
  Nanyang Technological University \\
  Singapore \\
  \texttt{xiakelin@ntu.edu.sg} \\
   \And
  Binh P. Nguyen \\
  School of Mathematics and Statistics \\
  Victoria University of Wellington \\
  Wellington, New Zealand \\
  \texttt{binh.p.nguyen@vuw.ac.nz} \\
}

\begin{document}
\maketitle

\begin{abstract}
Many molecular properties depend on 3D geometry, where non-covalent interactions, stereochemical effects, and medium- to long-range forces are determined by spatial distances and angles that cannot be uniquely captured by a 2D bond graph. Yet most 3D molecular graph neural networks still rely on globally fixed neighborhood heuristics, typically defined by distance cutoffs and maximum neighbor limits, to define local message-passing neighborhoods, leading to rigid, data-agnostic interaction budgets. We propose Multiscale Interaction Mixture of Experts (MI-MoE) to adapt interaction modeling across geometric regimes. Our contributions are threefold: (1) we introduce a distance-cutoff expert ensemble that explicitly captures short-, mid-, and long-range interactions without committing to a single cutoff; (2) we design a topological gating encoder that routes inputs to experts using filtration-based descriptors, including persistent homology features, summarizing how connectivity evolves across radii; and (3) we show that MI-MoE is a plug-in module that consistently improves multiple strong 3D molecular backbones across diverse molecular and polymer property prediction benchmark datasets, covering both regression and classification tasks. 
These results highlight topology-aware multiscale routing as an effective principle for 3D molecular graph learning.
\end{abstract}

\keywords{Graph Neural Networks \and Topological Deep Learning \and Mixture of Experts \and Molecular Representation}

\section{Introduction}

Predicting molecular properties is a central problem in drug discovery and materials science, enabling efficient screening and optimization of candidate molecules while substantially reducing experimental cost and time \cite{yang2019analyzing}. With the success of deep learning in natural language processing, computer vision, and graph learning, neural architectures have become a dominant paradigm for molecular property prediction \cite{gilmer2017neural}. 
Beyond SMILES strings and 2D molecular graphs, 3D molecular representations, commonly referred to as conformers, have attracted increasing attention as many molecular properties are governed by spatial atomic arrangements. In particular, non-covalent interactions, steric and stereochemical effects, and medium- to long-range interactions depend on interatomic distances and geometric configurations that cannot be uniquely inferred from a 2D bond graph~\cite{schutt2017schnet,Gasteiger2020Directional,fang2022geometry,10.1093/bib/bbac560}.

The development of 3D molecular graph neural networks (GNNs) has progressed from distance-based invariant models toward architectures that explicitly model geometric dependencies and equivariant representations. SchNet~\cite{schutt2017schnet} introduced continuous-filter convolutions based on interatomic distances, while DimeNet~\cite{Gasteiger2020Directional} and DimeNet++~\cite{gasteiger2020fast} improved expressiveness by incorporating angular information. Equivariant models such as PaiNN~\cite{schutt2021equivariant} and EGNN~\cite{satorras2021n} further preserve directional information by jointly modeling atomic features and coordinates, and more recent approaches, including GemNet~\cite{klicpera2021gemnet}, ViSNet~\cite{wang2024enhancing}, and MGNN~\cite{chang2025mgnn}, capture higher-order geometric interactions for improved molecular modeling.

In parallel, recent studies have shown that molecular representations extending beyond purely covalent-bond graphs, by incorporating non-covalent interactions or higher-order topological structures, can significantly enhance performance in molecular property prediction and related tasks \cite{wang2020topology,meng2021persistent,shen2023molecular,shen2024molecular}. Since intra-molecular interactions in conformers naturally involve both covalent and non-covalent effects, these findings motivate modeling molecular interactions across multiple spatial scales. However, most existing 3D GNNs still rely on globally fixed interaction heuristics, typically a single distance cutoff and a maximum number of neighbors, to construct message-passing neighborhoods \cite{schutt2017schnet,gasteiger2020fast,schutt2021equivariant,wang2024enhancing}. Such rigid, data-agnostic interaction budgets are unlikely to be optimal across molecules with diverse sizes, densities, and structural complexities, and they limit a model's ability to adapt its receptive field to interaction patterns.

To address this limitation, we propose a \emph{Multiscale Interaction Mixture of Experts (MI-MoE)} framework that explicitly models molecular interactions under multiple geometric regimes. MI-MoE constructs a set of experts operating on molecular graphs induced by distinct distance cutoffs, and introduces a topology-aware gating network that leverages filtration-based topological descriptors, including features derived from persistent homology, to adaptively weight experts. Intuitively, these descriptors summarize how molecular connectivity evolves as the interaction radius increases, providing a natural signal for multiscale routing. By integrating multiscale geometric representations with global topological cues, MI-MoE provides a principled mechanism for adaptively selecting and combining interaction scales in 3D molecular graphs. Extensive experiments on diverse molecular and polymer property prediction benchmarks show that MI-MoE consistently outperforms strong single-scale models, existing MoE-based approaches, and state-of-the-art baselines across both regression and classification tasks.

\noindent\textbf{Contributions.} Our main contributions are:
\begin{itemize}
    \item We introduce a multiscale MoE formulation for 3D molecular graphs using experts defined by multiple distance cutoffs, enabling interaction modeling across geometric regimes.
    \item We design a topology-aware gating mechanism based on filtration-derived descriptors to route and combine experts in an interaction-dependent manner.
    \item We demonstrate that MI-MoE is a plug-in module that improves multiple strong 3D backbones across diverse benchmarks and tasks.
\end{itemize}

\section{Related Work}
\subsection{3D Molecular Graph Neural Networks}

The development of 3D molecular graph neural networks has progressed from
distance-based invariant models toward architectures that explicitly encode
geometric dependencies and equivariant representations.
SchNet~\cite{schutt2017schnet} introduced continuous-filter convolutions based on
interatomic distances for rotationally invariant energy and force prediction,
while DimeNet~\cite{Gasteiger2020Directional} and its optimized variant
DimeNet++~\cite{gasteiger2020fast} enhanced expressiveness by incorporating
angular information through directional message passing.
To preserve directional features, Sch{\"u}tt~\textit{et~al}.~\cite{schutt2021equivariant}
proposed PaiNN, an equivariant message-passing framework with vector-valued
representations, and EGNN~\cite{satorras2021n} achieved equivariance by jointly
updating atomic coordinates and node embeddings without spherical harmonics.
More recent approaches further capture higher-order geometric interactions:
Klicpera~\textit{et~al}.~\cite{klicpera2021gemnet} introduced GemNet with
dihedral-aware two-hop message passing, while ViSNet~\cite{wang2024enhancing} and
MGNN~\cite{chang2025mgnn} employed vector–scalar interaction modeling and
moment-based representations to achieve strong performance in molecular
potential prediction and dynamic simulations.

\subsection{Mixture of Experts in Graph Learning}

Mixture of Experts (MoE) models~\cite{jacobs1991adaptive,shazeer2017} combine
multiple specialized experts and have achieved notable success in large-scale
learning systems, while their application to graph learning remains relatively
limited.
To address depth sensitivity in graph representations, Yao~\textit{et~al}.~\cite{yao2025moe}
proposed DA-MoE, which employs GNN experts with varying depths and a
structure-aware gating mechanism, and Wang~\textit{et~al}.~\cite{wang2023graph}
introduced GMoE, which utilizes experts with different hop ranges to capture both
short- and long-range interactions.
Beyond aggregation depth, MoE frameworks have been extended to handle spectral
and geometric heterogeneity: NodeMoE~\cite{han2024node} dynamically selects
experts with distinct spectral filters to accommodate homophilic and
heterophilic patterns, while GraphMoRE~\cite{guo2025graphmore} addresses
topological diversity by combining Riemannian experts operating in Euclidean,
hyperbolic, and spherical spaces.
In the context of molecular property prediction, TopExpert~\cite{kim2023learning}
assigns topology-specific experts via clustering-based gating, and
GraphMETRO~\cite{wu2024graphmetro} improves robustness to distribution shifts by
decomposing shifts into interpretable components and aligning multiple experts to
learn invariant graph representations.

\subsection{Topological Molecular Learning}

Recent advances in molecular representation learning have increasingly integrated
topological information with geometric deep learning to overcome the limitations
of traditional 2D graph and sequence-based representations.
To recover stereochemical and spatial information lost in sequence models,
Chen~\textit{et~al}.~\cite{chen2021algebraic} proposed AGBT, which fuses
algebraic graph fingerprints derived from multiscale weighted colored graphs to
encode 3D molecular structure, while Rong~\textit{et~al}.~\cite{rong2025topological}
introduced a topological fusion network that extracts 1D and 2D simplices
representing covalent bonds and functional groups.
Moving beyond pairwise graphs, HL-HGAT~\cite{huang2025hl} models molecules as
simplicial complexes using Hodge-Laplacian operators, a perspective further
extended by Mol-TDL~\cite{shen2024molecular} through filtration-based simplicial
complexes and topological contrastive learning.
Challenging the reliance on covalent-bond graphs, Mol-GDL~\cite{shen2023molecular}
demonstrated that non-covalent distance-based representations can better capture
long-range interactions.
Complementing these approaches, Dong~\textit{et~al}.~\cite{dong2025exploring}
employed topological data analysis within a multi-objective optimization framework
to automatically select complementary molecular representations across
fingerprints, graphs, and 3D conformations.

\paragraph{Our position.}
While recent advances in 3D molecular graph neural networks have substantially improved geometric expressiveness under fixed interaction scales, and mixture of experts models have addressed graph heterogeneity through architectural depth, hop range, or latent geometry, these paradigms remain largely decoupled from topological reasoning. Existing MoE approaches such as DA-MoE and GMoE route graphs based on structural depth or aggregation scope, without explicitly modeling multiscale molecular interactions, whereas topology-aware molecular learning methods typically incorporate topological descriptors as auxiliary features rather than as decision-making signals. In contrast, MI-MoE unifies multiscale 3D geometric learning, filtration-based topological analysis, and mixture of experts modeling within a single end-to-end framework. MI-MoE defines experts over distinct physically meaningful interaction cut-off radii and employs a topology-aware gating network that leverages the evolution of molecular connectivity and higher-order structure across radii to dynamically route each molecule to appropriate experts. Unlike prior 3D GNNs that operate at a single fixed scale, MI-MoE enables adaptive multiscale reasoning; unlike existing MoE-based graph models, expert specialization is grounded in molecular interaction physics rather than architectural variation; and unlike prior topological molecular learning methods, topological information in MI-MoE directly governs expert routing rather than being fused post hoc. This design positions MI-MoE as a principled framework for capturing heterogeneous molecular interaction regimes through topology-guided expert selection.

\section{Method}

\begin{figure*}[t]
\centering
\includegraphics[width=1\textwidth]{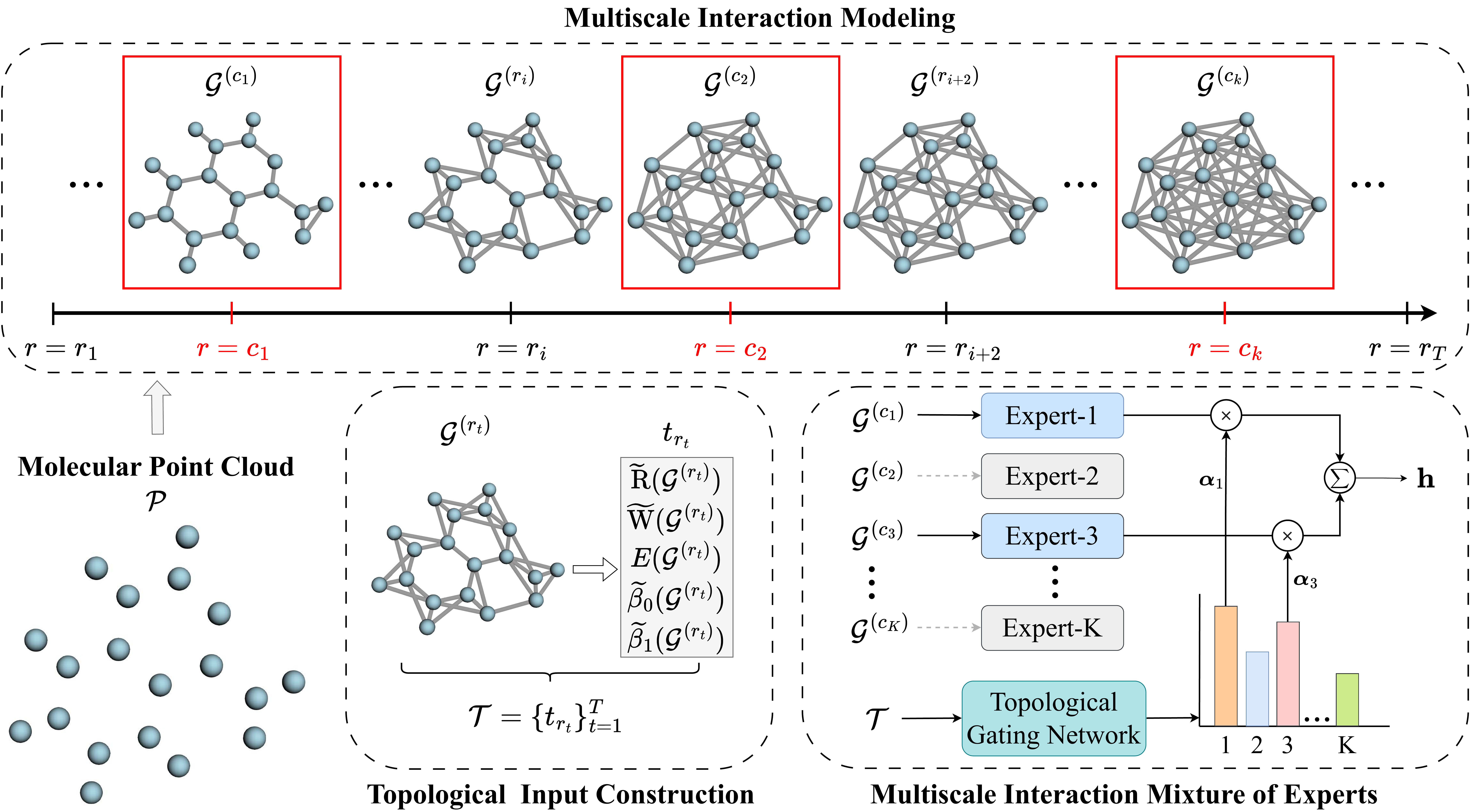}
\caption{Overview of the MI-MoE framework. Starting from a molecular point cloud $\mathcal{P}$, a distance-based filtration induces a family of multiscale interaction graphs. From this filtration, two graph sets are derived: (i) a sparse set of cutoff-specific graphs $\{\mathcal{G}^{(c_k)}\}_{k=1}^{K}$ used as inputs to the expert models, and (ii) a denser set of graphs $\{\mathcal{G}^{(r_t)}\}_{t=1}^{T}$ used to compute topological descriptors $\widetilde{\mathrm{R}}(\mathcal{G}^{(r_t)}),\widetilde{\mathrm{W}}(\mathcal{G}^{(r_t)}),E(\mathcal{G}^{(r_t)}),\widetilde{\beta}_0(\mathcal{G}^{(r_t)}),\widetilde{\beta}_1(\mathcal{G}^{(r_t)})$ that characterize the evolution of molecular connectivity across geometric scales for routing. These descriptors are fed into a topological gating network that produces interaction-dependent routing weights $\boldsymbol{\alpha}$ over experts. Finally, the selected expert outputs are aggregated according to $\boldsymbol{\alpha}$ to produce the final molecular representation $\mathbf{h}$.}
\label{fig:MI_MOE_Framework}
\vspace{-.3cm}
\end{figure*}

\textit{Notation.}
A molecule is modeled as a point cloud $\mathcal{P} = (\mathcal{V}, \mathbf{R})$,
where $\mathcal{V}$ denotes the set of atoms and
$\mathbf{R} \in \mathbb{R}^{|\mathcal{V}| \times 3}$ denotes their Cartesian coordinates.
Each atom $v_i \in \mathcal{V}$ is associated with an initial feature vector
$\mathbf{h}_i \in \mathbb{R}^{d}$, and $\mathbf{r}_i=(x_i,y_i,z_i)$ denotes its position.
Given only positions and features, we do not assume explicit connectivity \emph{a priori}.
Instead, we induce an interaction graph
$\mathcal{G}=(\mathcal{V},\mathcal{E},\mathbf{R})$
by a geometric rule on atom pairs, and associate each edge $(i,j)\in\mathcal{E}$
with an edge feature vector $\mathbf{e}_{ij}\in\mathbb{R}^{d_e}$ derived from geometry
(e.g., distance- or angle-related attributes).
Unless stated otherwise, interaction graphs are undirected and exclude self-loops.

\paragraph{Overview.}
Given $\mathcal{P}$, we construct a distance-based filtration and derive two graph families:
(i) a sparse set of cutoff graphs $\{\mathcal{G}^{(c_k)}\}_{k=1}^{K}$ for expert backbones and
(ii) a denser sequence $\{\mathcal{G}^{(r_t)}\}_{t=1}^{T}$ for computing topology descriptors that
summarize how connectivity evolves across radii.
A topological encoder maps the descriptor trajectory to routing logits, we apply Top-$k$ sparsification
and softmax to obtain expert weights, and then aggregate selected expert representations into a graph-level
embedding used for prediction. Figure \ref{fig:MI_MOE_Framework} illustrates the overview of our proposed MI-MoE framework.

\subsection{Multiscale Graph Construction}
\label{subsec:multi_scale_construction}

Given a molecule represented as a 3D point cloud $\mathcal{P}=(\mathcal{V},\mathbf{R})$,
we induce interaction graphs through a distance-based rule. For a cutoff radius $r$, we define
\begin{align}
\mathcal{G}^{(r)} &= (\mathcal{V}, \mathcal{E}^{(r)}, \mathbf{R}), \\
\mathcal{E}^{(r)} &= \left\{ (i,j)\in\mathcal{V}\times\mathcal{V},\, i\neq j \;\middle|\;
\lVert \mathbf{r}_i - \mathbf{r}_j \rVert_2 \le r \right\}.
\end{align}
By increasing $r$, this construction induces a nested sequence
\begin{equation}
\mathcal{G}^{(r_1)} \subseteq \mathcal{G}^{(r_2)} \subseteq \cdots \subseteq \mathcal{G}^{(r_M)},
\quad r_1 < r_2 < \cdots < r_M,
\end{equation}
which forms a filtration in the sense of topological data analysis and provides a principled mechanism
for capturing molecular interactions across multiple geometric scales \cite{shen2024molecular}.

\subsection{Multiscale Interaction Modeling}
\label{subsec:multiscale_modeling}

At the atomic level, covalent connectivity can be recovered by connecting atom pairs within a short
distance threshold of approximately $2\,\text{\AA}$ \cite{shen2023molecular}. Motivated by this observation,
we adopt $2\,\text{\AA}$ as the smallest radius in our filtration.

\paragraph{Expert cutoffs.}
From the filtration, we select a finite set of interaction radii
\begin{equation}
\{c_k\}_{k=1}^{K}=\{2.0,\,2.5,\,3.0,\,3.5,\,4.0\}\,\text{\AA},
\end{equation}
to define $K$ interaction-scale experts. Each $c_k$ induces a graph $\mathcal{G}^{(c_k)}$ on which the
$k$-th expert operates. These radii progressively expand the interaction neighborhood and cover both
covalent and non-covalent effects.
For compatibility with standard 3D GNN implementations, one may optionally impose a maximum-neighbor
cap when constructing expert graphs (e.g., keeping the closest neighbors within the cutoff); MI-MoE is
agnostic to this choice.

\paragraph{Dense radii for routing.}
To enable topology-aware routing, we derive routing inputs from the same filtration using a denser
discretization of the distance axis. We consider radii
$\{r_t\}_{t=1}^{T}$ that densely sample an extended interval
$[c_1-w/2,\;c_K+w/2]$, where $w>0$ is a window hyperparameter, and satisfy
$\{c_k\}_{k=1}^{K} \subseteq \{r_t\}_{t=1}^{T}$.
In practice, we discretize the interval with a fixed step size $\Delta r$,
yielding $T=\left\lfloor \frac{w+(c_K-c_1)}{\Delta r}\right\rfloor+1$ radii.
The resulting dense filtration graphs $\{\mathcal{G}^{(r_t)}\}_{t=1}^{T}$ are used exclusively for
topology computation. This design decouples sparse expert execution from dense structural analysis,
so routing can exploit how molecular connectivity evolves between and around expert cutoffs, rather
than relying on isolated cutoff values.

\subsection{Expert Networks}
\label{subsec:expert_models}

To capture interactions under different geometric regimes, we employ $K$ expert models
$\{E_k(\cdot)\}_{k=1}^{K}$, each operating on its corresponding cutoff graph $\mathcal{G}^{(c_k)}$:
\begin{equation}
\label{eq:multiscale_graphs}
\left\{
\mathcal{G}^{(c_k)} = (\mathcal{V}, \mathcal{E}^{(c_k)}, \mathbf{R});
\quad k = 1, \dots, K
\right\}.
\end{equation}
Each expert is instantiated as an independent 3D GNN backbone that shares the same architecture across experts but with separate parameters and operates on its corresponding interaction graph $\mathcal{G}^{(c_k)}$, allowing specialization to a particular interaction scale.

At layer $l$ of expert $E_k(\cdot)$, each node $v_i$ maintains invariant features
$\mathbf{h}_i^{(l)}\in\mathbb{R}^{d}$ and equivariant geometric features $\mathbf{x}_i^{(l)}$, representing either atomic coordinates or learned vector features. For each edge $(i,j)\in\mathcal{E}^{(c_k)}$,
the message from $v_j$ to $v_i$ is computed as
\begin{equation}
\mathbf{m}_{i,j}^{(l)} =
\texttt{MSG}^{(l)}\!\left(
\mathbf{h}_i^{(l-1)}, \mathbf{h}_j^{(l-1)},
\mathbf{x}_i^{(l-1)}, \mathbf{x}_j^{(l-1)},
\mathbf{e}_{ij}
\right),
\end{equation}
where $\texttt{MSG}^{(l)}(\cdot)$ is invariant or equivariant with respect to Euclidean transformations.
Messages are aggregated and used to update node states by $\texttt{AGG}^{(l)}(\cdot)$ and $\texttt{UPD}^{(l)}(\cdot)$, respectively:
\begin{align}
\mathbf{m}_i^{(l)} &=
\texttt{AGG}^{(l)}\left( \left \{ \mathbf{m}_{i,j}^{(l)} \mid j \in \mathcal{N}_{c_k}(i) \right \} \right),\\
\left(\mathbf{h}_i^{(l)}, \mathbf{x}_i^{(l)}\right)&=
\texttt{UPD}^{(l)}\left(
\mathbf{h}_i^{(l-1)}, \mathbf{x}_i^{(l-1)}, \mathbf{m}_i^{(l)}
\right),
\end{align}
where $\mathcal{N}_{c_k}(i)$ denotes neighbors under cutoff $c_k$.
After $L$ layers, a graph-level embedding is obtained via permutation-invariant readout $\texttt{READOUT}(\cdot)$:
\begin{equation}
\mathbf{h}_{\mathcal{G}^{(c_k)}}=
\texttt{READOUT}\left(\{\mathbf{h}_i^{(L)} \mid i \in \mathcal{V}\}\right).
\end{equation}
Experts with smaller cutoffs emphasize local, dense interactions, while those with larger cutoffs capture more extended and sparser geometric dependencies.

\subsection{Topological Gating Mechanism}
\label{subsec:topo_gate_networks}

\subsubsection{Topological descriptors on the dense filtration}
\label{subsec:topo_input}

We compute a compact set of normalized topological descriptors for each dense-filtration graph
$\mathcal{G}^{(r_t)}=(\mathcal{V},\mathcal{E}^{(r_t)},\mathbf{R})$.
Inspired by recent results showing the effectiveness of topology-aware graph features
\cite{wesolowski2025fast}, we use degree- and distance-based indices, including Randi\'{c}~\cite{randic1975characterization}, Wiener~\cite{wiener1947structural}, and global efficiency~\cite{latora2001efficient}, together with topological features derived from persistent homology, namely the Betti curves associated with the $0$- and $1$-D homology groups.
Although the Randi\'{c} and Wiener indices are widely adopted molecular descriptors, their raw values depend on graph size and interaction density. Normalization is therefore required to obtain stable and comparable signals across different graphs defined by different interaction cutoffs, densities, and molecular sizes.

\paragraph{Topological motivation for scale selection.}
From a topological perspective, distance-based filtrations capture how molecular connectivity evolves as the interaction radius increases. Abrupt changes in Betti numbers or global graph indices indicate structural transitions, while stable regions correspond to radii at which local and non-local interactions are balanced. MI-MoE exploits this stability signal to guide expert selection, rather than assuming a single fixed interaction.

\paragraph{Randi\'{c} index.}
Let $\deg_t(v)$ be the degree of node $v$ in $\mathcal{G}^{(r_t)}$ and $n=|\mathcal{V}|$. The original and normalized versions of Randi\'{c} index are formally defined as follows:
\begin{align}
\label{eq:randic_norm}
\mathrm{R}(\mathcal{G}^{(r_t)}) &=
\sum_{(u,v)\in\mathcal{E}^{(r_t)}}
\frac{1}{\sqrt{\deg_t(u)\deg_t(v)}}, \\
\widetilde{\mathrm{R}}(\mathcal{G}^{(r_t)}) &=
\frac{\mathrm{R}(\mathcal{G}^{(r_t)})}{n/2}.
\end{align}
This captures degree correlations and local branching structure while remaining bounded across varying
graph sizes.

\paragraph{Wiener index.}
Let $\mathcal{C}_t$ be the set of connected components of $\mathcal{G}^{(r_t)}$.
For each component $C\in\mathcal{C}_t$, let $d_C(u,v)$ be the shortest-path distance within $C$. The original Wiener index is defined as:
\begin{equation}
\label{eq:wiener_raw}
\mathrm{W}(\mathcal{G}^{(r_t)}) =
\sum_{C\in\mathcal{C}_t}
\sum_{\substack{u,v\in C,\, u<v}} d_C(u,v).
\end{equation}
For a connected graph with $n$ nodes, extremal bounds are
$\mathrm{W}_{\min}(n)=\frac{n(n-1)}{2}$ (complete graph) and
$\mathrm{W}_{\max}(n)=\frac{n^3-n}{6}$ (path graph).
We define the normalized Wiener index:
\begin{equation}
\label{eq:wiener_norm}
\widetilde{\mathrm{W}}(\mathcal{G}^{(r_t)})=
\frac{\mathrm{W}(\mathcal{G}^{(r_t)})-\mathrm{W}_{\min}(n)}
{\mathrm{W}_{\max}(n)-\mathrm{W}_{\min}(n)}.
\end{equation}

This normalization ensures bounded by the graph size while preserving the relative ordering of global compactness.

\paragraph{Global efficiency.}
We incorporate the global efficiency to characterize the overall communicability of an interaction graph.
The global efficiency is defined as:
\begin{equation}
E(\mathcal{G}^{(r_t)})=\frac{1}{n(n-1)}
\sum_{\substack{u,v\in\mathcal{V},\, u\neq v}}
\frac{1}{d(u,v)},
\end{equation}
where $d(u,v)$ is the shortest-path distance, and disconnected pairs contribute zero.
It complements Wiener by emphasizing short-range connectivity and communication.

\paragraph{Betti curves from persistent homology.}
We compute persistent homology over the filtration induced by $\{\mathcal{G}^{(r_t)}\}_{t=1}^{T}$.
Let $\mathcal{D}_p=\{(b_i,d_i)\}$ be the persistence diagram for $p$-dimensional homology, where $p=0$ for components and $p=1$ for loops. Following \cite{chung2022persistence}, we define normalized
feature weights
\begin{equation}
\label{eq:betti_weight}
\psi(b_i,d_i)=
\frac{\phi(b_i,d_i)}
{\sum_{(b_j,d_j)\in\mathcal{D}_p}|\phi(b_j,d_j)|},
\end{equation}
where $\phi$ is a weighting function (e.g., $\phi\equiv 1$). The normalized Betti curve at radius $r_t$ is
\begin{equation}
\label{eq:betti_curve}
\widetilde{\beta}_p(\mathcal{G}^{(r_t)})=
\sum_{(b_i,d_i)\in\mathcal{D}_p}
\psi(b_i,d_i)\,\mathbb{I}[\,b_i\le r_t < d_i\,].
\end{equation}
This yields bounded, stable summaries of topological evolution across radii~\cite{chung2022persistence}.

\subsubsection{Topological trajectory and routing}
\label{subsec:topo_gating}

For each interaction graph $\mathcal{G}^{(r_t)}$, we form a descriptor vector
\begin{equation}
\mathbf{t}_{r_t}
=
\left\{ f_i\!\left(\mathcal{G}^{(r_t)}\right) \right\}_{i=1}^{5},
\quad
f =
\big(
\widetilde{\mathrm{R}},
\widetilde{\mathrm{W}},
E,
\widetilde{\beta}_0,
\widetilde{\beta}_1
\big),
\end{equation}
collecting over the dense filtration yields a topological trajectory $\mathcal{T}=\{\mathbf{t}_{r_t}\}_{t=1}^{T}$, equivalently a matrix $X_{\text{topo}}\in\mathbb{R}^{T\times 5}$, 
which provides a compact yet expressive characterization of the evolution of molecular connectivity, distance-based structure, and higher-order topology as the interaction radius increases. 
Although $\widetilde{\beta}_0$ typically stabilizes once the interaction graph becomes fully connected (often shortly after 2\,\text{\AA}), it remains informative by encoding variations in the radius at which connectivity emerges and by reflecting early-stage component merging across molecules.
We use a learnable  
topological encoder $f_{\mathrm{topo}}(\cdot)$ to map $\mathcal{T}$ to routing logits:
\begin{equation}
\boldsymbol{\alpha}^{\mathrm{raw}}=f_{\mathrm{topo}}(\mathcal{T})\in\mathbb{R}^{K},
\end{equation}
where each entry corresponds to an unnormalized importance score for one expert.
By default, $f_{\mathrm{topo}}(\cdot)$ can be implemented as a multilayer perceptron (MLP) on the flattened $X_{\text{topo}}$;
alternatively, attention-based encoders as 
Transformers can model dependencies across radii.
To encourage sparse expert activation, we retain only the Top-$k$ logits and suppress others:
\begin{equation}
\widetilde{\boldsymbol{\alpha}}=\mathrm{TopK}\!\left(\boldsymbol{\alpha}^{\mathrm{raw}},k\right),
\end{equation}
where $\mathrm{TopK}(\cdot)$ preserves the $k$ largest elements and sets the rest to $-\infty$.
Final expert weights are obtained by softmax:
\begin{equation}
\boldsymbol{\alpha}=\mathrm{softmax}(\widetilde{\boldsymbol{\alpha}}).
\end{equation}
This formulation enables sparse, molecule-dependent expert selection while
maintaining differentiability during training.

\subsection{Mixture of Experts over Multiscale Graphs}
\label{subsec:moe_experts}

Given expert graph embeddings $\{\mathbf{h}_{\mathcal{G}^{(c_k)}}\}_{k=1}^{K}$ and weights
$\boldsymbol{\alpha}=(\alpha_1,\dots,\alpha_K)$,
we model molecular interactions across multiple geometric regimes to construct the final molecular representation:
\begin{equation}
\label{eq:moe_aggregate}
\mathbf{h}=\sum_{k=1}^{K}\alpha_k\,\mathbf{h}_{\mathcal{G}^{(c_k)}}.
\end{equation}
This aggregation adaptively emphasizes scale-specific representations while allowing contributions from multiple scales when beneficial.

\begin{algorithm}[t]
\caption{MI-MoE forward pass}
\label{alg:mimoe}
\begin{algorithmic}[1]
\Require Point cloud $\mathcal{P}=(\mathcal{V},\mathbf{R})$, expert cutoffs $\{c_k\}_{k=1}^{K}$, dense radii $\{r_t\}_{t=1}^{T}$, Top-$k$ value $k$
\Ensure Molecular representation $\mathbf{h}$
\State Construct expert graphs $\{\mathcal{G}^{(c_k)}\}_{k=1}^{K}$ and dense graphs $\{\mathcal{G}^{(r_t)}\}_{t=1}^{T}$
\State Compute descriptors $\mathbf{t}_{r_t}$ for each $\mathcal{G}^{(r_t)}$ and form $\mathcal{T}=\{\mathbf{t}_{r_t}\}_{t=1}^{T}$
\State $\boldsymbol{\alpha}^{\mathrm{raw}} \gets f_{\mathrm{topo}}(\mathcal{T})$
\State $\boldsymbol{\alpha} \gets \mathrm{softmax}\!\left(\mathrm{TopK}(\boldsymbol{\alpha}^{\mathrm{raw}},k)\right)$
\State For each expert $k$, compute $\mathbf{h}_{\mathcal{G}^{(c_k)}} \gets E_k(\mathcal{G}^{(c_k)})$
\State $\mathbf{h} \gets \sum_{k=1}^{K} \alpha_k\,\mathbf{h}_{\mathcal{G}^{(c_k)}}$
\State \Return $\mathbf{h}$
\end{algorithmic}
\end{algorithm}

\subsection{Balanced Expert Utilization Loss}
\label{subsec:balanced_loss}

MoE models may suffer from \emph{expert imbalance}, where the gate over-selects a subset of experts,
leading to inefficient capacity usage. We add an auxiliary balancing loss that regularizes both
(i) the magnitude of routing logits and (ii) the selection probabilities under sparse Top-$k$ routing,
following common MoE balancing principles \cite{wang2023graph,yao2025moe}.

\paragraph{Expert score balance.}
For a mini-batch of size $B$, let $\boldsymbol{\alpha}^{(b)}\in\mathbb{R}^{K}$ denote the final expert weights for sample $b$.
We define the accumulated activation vector
\begin{equation}
\mathbf{s}=\sum_{b=1}^{B}\boldsymbol{\alpha}^{(b)} \in \mathbb{R}^K,
\end{equation}
and penalize deviations from its mean to discourage systematically large logits for a subset of experts:
\begin{equation}
\mathcal{L}_{\text{score}}=
\frac{1}{K}\sum_{i=1}^{K}\frac{(s_i-\bar{s})^2}{\bar{s}^2+\epsilon},
\quad
\bar{s}=\frac{1}{K}\sum_{i=1}^{K}s_i.
\end{equation}

\paragraph{Expert selection balance.}
Balancing logits alone does not fully prevent imbalanced utilization under sparse routing.
Let $P(b,i) = \alpha_i^{(b)}$ denote the selection probability of expert $i$ for sample $b$, with $P(b,i)=0$ for experts that are not selected.
We accumulate expert importance over the batch:
\begin{equation}
\mathcal{P}_i=\sum_{b=1}^{B}P(b,i),
\qquad
\bar{\mathcal{P}}=\frac{1}{K}\sum_{i=1}^{K}\mathcal{P}_i,
\end{equation}
and minimize its dispersion:
\begin{equation}
\mathcal{L}_{\text{load}}=
\frac{1}{K}\sum_{i=1}^{K}\frac{(\mathcal{P}_i-\bar{\mathcal{P}})^2}{\bar{\mathcal{P}}^2+\epsilon}.
\end{equation}
This explicitly penalizes experts that are over- or under-selected, promoting balanced load under Top-$k$
routing.

\subsection{Overall Training Objective}
\label{subsec:objective}

The overall training objective combines the task loss with balancing regularization:
\begin{equation}
\mathcal{L}_{\text{total}}=
\mathcal{L}_{\text{task}}
+\lambda_1\mathcal{L}_{\text{score}}
+\lambda_2\mathcal{L}_{\text{load}},
\end{equation}
where $\mathcal{L}_{\text{task}}$ is the primary task loss, and $\lambda_1,\lambda_2$ control the
strength of the balancing terms. In our experiments, we set $\lambda_1=\lambda_2=\lambda$.


\section{Experiments}
\label{sec:experiments}

\subsection{Backbone Layers}
\label{subsec:backbone_layers}

The proposed MI-MoE framework is designed as a plug-and-play module and can be integrated with different 3D GNN architectures serving as expert models. In our experiments, we evaluate a set of representative invariant backbone architectures, including SchNet~\cite{schutt2017schnet}, modified invariant variants of the Graph Isomorphism Network (GIN)~\cite{xu2018how} and Graph Transformer Network (GTN)~\cite{shi2021masked}, as well as DimeNet++~\cite{gasteiger2020fast} and ViSNet~\cite{wang2024enhancing}. We further consider equivariant architectures, including EGNN~\cite{satorras2021n} and PaiNN~\cite{schutt2021equivariant}, which explicitly model geometric transformations through equivariant message passing. The depth of each 3D GNN expert is set to be shallower than that of the corresponding standalone baseline. Detailed configurations of all expert models are provided in the Appendix.

\subsection{3D Molecular Representation Generation}

Three-dimensional molecular representations are generated from SMILES strings using RDKit~\cite{landrum2013rdkit}, which employs a distance geometry-based embedding with the Experimental-Torsion Knowledge Distance Geometry~\cite{riniker2015better} scheme to incorporate torsional preferences. The resulting structures are further refined through molecular mechanics energy minimization, using the Merck Molecular Force Field algorithm~\cite{halgren1996merck} when available, to alleviate steric clashes and improve geometric consistency. The optimized atomic coordinates are subsequently used as 3D molecular point clouds for the MI-MoE framework.

\subsection{Implementation Settings}
\label{subsec:impl_settings}

In all experiments, the hidden layer dimension is set to 128. Training hyperparameters are tuned by selecting batch sizes from $\{32, 64\}$, learning rates from $\{10^{-4}, 10^{-3}\}$, and dropout rates from $\{0, 0.5\}$. We employ a cosine annealing learning rate schedule with an initial warm-up phase. Models are trained for up to 120 epochs, with early stopping applied if the validation performance does not improve for 30 consecutive epochs. All experiments are conducted on a server equipped with four NVIDIA A40 GPUs (48GB each).

\subsection{Molecular Property Prediction Tasks}
\subsubsection{Datasets}
\label{subsec:datasets}

We evaluate MI-MoE on eight benchmark datasets from MoleculeNet~\cite{wu2018moleculenet}, covering both regression and classification tasks: FreeSolv, ESOL, Lipophilicity, BACE, BBBP, ClinTox, SIDER, and Tox21.
We follow the data curation and scaffold splitting protocol described in~\cite{nguyen2024smiles}.
Additional dataset details are provided in the Appendix.

\subsubsection{Baselines}
\label{subsec:baselines}

We compare against representative 2D GNN-based molecular property prediction
models, including AttentiveFP~\cite{xiong2019pushing}, 
D-MPNN~\cite{yang2019analyzing},
HiGNN~\cite{zhu2022hignn}, 
ResGAT~\cite{nguyen2024resgat},
HiMol~\cite{zang2023hierarchical}, and MolCLR~\cite{wang2022molecular}.
We additionally include Transformer-based models operating on SMILES, including
ChemBERTa~\cite{chithrananda2020chemberta}, and Uni-Mol~\cite{zhou2023unimol},
which explicitly incorporates 3D geometric information.
Finally, we compare with MoE-based baselines, including TopExpert~\cite{kim2023learning}
and DA-MoE~\cite{yao2025moe}. Additional details are in the Appendix.

\subsubsection{Performance Evaluation}
\label{subsec:perf_eval}

Tables~\ref{tab:3d_baselines_moe} and~\ref{tab:result_all} summarize the performance of MI-MoE on MoleculeNet regression and classification benchmarks. Across the evaluated 3D GNN backbones, integrating MI-MoE leads to performance improvements in most cases for both regression and classification tasks. For instance, when applied to SchNet, MI-MoE reduces the average RMSE from 1.23 to 1.06 and increases the average ROC-AUC from 69.70\% to 81.29\%. Equivariant architectures also benefit from MI-MoE: for PaiNN, the average RMSE improves from 1.09 to 0.97, while the average ROC-AUC increases from 72.77\% to 79.92\%, suggesting that multiscale expert routing is complementary to equivariant message passing. For already strong backbones such as ViSNet, MI-MoE achieves comparable overall performance, with a modest improvement in average RMSE from 0.98 to 0.97 and competitive average ROC-AUC (79.97\% vs.\ 79.66\%). Compared with state-of-the-art models such as Uni-Mol and TopExpert, MI-MoE demonstrates competitive or superior performance, indicating the effectiveness of the proposed framework. Overall, these results suggest that topology-aware multiscale expert routing enhances molecular representation learning by enabling adaptive interaction modeling across geometric scales.

\begin{table*}[!ht]
\centering
\caption{Average RMSE and ROC-AUC (\%) results (with standard deviation) on MoleculeNet regression and classification tasks, comparing original 3D GNN baselines against those enhanced by MI-MoE. Best results are in bold, and OOM denotes Out-of-Memory.}
\begin{adjustbox}{width=1\textwidth}
\begin{tabular}{lccc|c|ccccc|c}
\toprule
& \multicolumn{4}{c|}{Regression (RMSE $\downarrow$)}  &\multicolumn{6}{c}{Classification (ROC-AUC (\%) $\uparrow$)}\\
\cmidrule(lr){2-5} \cmidrule(lr){6-11}
Model & FreeSolv & ESOL & Lipophilicity & Avg. & BACE & BBBP & SIDER & Tox21 & ClinTox & Avg. \\
\midrule
GIN & 4.23 (1.19) & 1.09 (0.08) & 1.04 (0.13) & 2.12 &
76.24 (4.2) & 82.09 (6.0) & 60.23 (3.1) & 77.89 (3.0) & 64.79 (8.8) & 72.25\\
+ MI-MoE & \textbf{2.33 (0.38)} & \textbf{0.94 (0.15)} & \textbf{0.70 (0.04)} & \textbf{1.32} &
\textbf{85.15 (1.8)} & \textbf{90.46 (3.5)} & \textbf{60.72 (1.4)} & \textbf{80.39 (2.3)} & \textbf{89.04 (8.1)} & \textbf{81.15} \\
\midrule
GTN & \textbf{1.75 (0.50)} & 0.84 (0.07) & 0.86 (0.06) & 1.15 &
79.75 (5.9) & 84.79 (6.0) & 59.96 (2.7) & 78.53 (1.8) & 65.17 (10.0) & 73.64 \\
+ MI-MoE & \textbf{1.75 (0.22)} & \textbf{0.83 (0.14)} & \textbf{0.68 (0.06)} & \textbf{1.09} &
\textbf{82.76 (4.6)} & \textbf{89.47 (6.1)} & \textbf{61.84 (3.3)} & \textbf{79.79 (1.9)} & \textbf{90.77 (4.5)} & \textbf{80.93} \\
\midrule
SchNet & 1.91 (0.21) & 0.97 (0.14) & 0.81 (0.06) & 1.23 &
74.38 (4.9) & 78.36 (16.3) & 59.10 (3.1) & 77.59 (2.5) & 59.07 (12.3) & 69.70\\
+ MI-MoE & \textbf{1.76 (0.37)} & \textbf{0.78 (0.13)} & \textbf{0.63 (0.04)} & \textbf{1.06} &
\textbf{82.19 (2.7)} & \textbf{89.69 (4.9)} & \textbf{61.42 (2.4)} & \textbf{80.43 (2.0)} & \textbf{92.71 (3.9)} & \textbf{81.29} \\
\midrule
DimeNet++ & 1.97 (0.59) & 0.79 (0.13) & 0.65 (0.03) & 1.14 &
\textbf{82.27 (4.5)} & 88.32 (4.2) & 61.38 (5.0) & 78.68 (2.8) & 83.05 (12.2) & 78.74\\
+ MI-MoE & \textbf{1.80 (0.39)} & \textbf{0.71 (0.10)} & \textbf{0.61 (0.02)} & \textbf{1.04} &
81.57 (3.4) & \textbf{88.35 (6.4)} & OOM & \textbf{80.95 (2.0)} & \textbf{83.63 (19.3)} & \textemdash \\
\midrule
ViSNet & \textbf{1.52 (0.31)} & 0.76 (0.10) & 0.66 (0.03) & 0.98 &
\textbf{85.59 (2.8)} & \textbf{89.41 (2.1)} & \textbf{61.13 (4.5)} & 77.44 (2.5) & 86.28 (6.6) & \textbf{79.97} \\
+ MI-MoE & 1.58 (0.24) & \textbf{0.72 (0.10)} & \textbf{0.62 (0.04)} & \textbf{0.97} &
82.37 (4.0) & 88.68 (5.3) & 59.72 (2.2) & \textbf{78.14 (3.2)} & \textbf{89.39 (4.8)} & 79.66\\
\midrule
EGNN & 3.04 (0.81) & 1.43 (0.38) & 0.81 (0.12) & 1.76 &
58.29 (9.4) & 79.92 (2.4) & 57.85 (3.5) & 77.91 (1.2) & 62.32 (8.4) & 67.26\\
+ MI-MoE & \textbf{1.64 (0.33)} & \textbf{0.78 (0.06)} & \textbf{0.61 (0.04)} & \textbf{1.01} &
\textbf{78.60 (4.8)} & \textbf{89.49 (3.8)} & \textbf{60.41 (3.4)} & \textbf{81.06 (1.9)} & \textbf{90.32 (6.2)} & \textbf{79.98} \\
\midrule
PaiNN & 1.59 (0.41) & 0.99 (0.17) & 0.69 (0.04) & 1.09 &
77.69 (4.7) & 86.98 (3.0) & 55.65 (2.3) & 78.25 (3.1) & 65.26 (6.1) & 72.77\\
+ MI-MoE & \textbf{1.52 (0.28)} & \textbf{0.78 (0.11)} & \textbf{0.61 (0.02)} & \textbf{0.97} &
\textbf{81.31 (5.2)} & \textbf{87.95 (2.9)} & \textbf{58.78 (2.3)} & \textbf{80.69 (2.7)} & \textbf{90.85 (4.6)} & \textbf{79.92} \\
\bottomrule
\end{tabular}
\end{adjustbox}
\label{tab:3d_baselines_moe}
\vspace{-.3cm}
\end{table*}

\begin{table*}[!ht]
\centering
\caption{Average RMSE and ROC-AUC (\%) results (including standard deviations) on MoleculeNet regression and classification tasks. Best results are in bold; second-best are underlined. TopExpert does not support regression tasks by default.}
\begin{adjustbox}{width=1\textwidth}
\begin{tabular}{l|ccc|c|ccccc|c|}
\toprule
& \multicolumn{4}{|c|}{Regression (RMSE $\downarrow$)}  &\multicolumn{6}{|c|}{Classification (ROC-AUC (\%) $\uparrow$)}\\
\cmidrule(lr){2-5} \cmidrule(lr){6-11}
Model & FreeSolv  & ESOL & Lipophilicity & Avg. & BACE & BBBP & SIDER & Tox21 & ClinTox & Avg. \\
\midrule
\textbf{Previous studies} &  &  & & & & & & & \\
AttentiveFP & 4.99 (0.47) & 1.74 (0.36) & 1.11 (0.07) & 2.61 &
81.74 (5.1) & 87.86 (5.6) & 58.77 (3.6) & \textbf{82.23 (2.2)} & 75.65 (16.0) & 77.25 \\
D-MPNN & 1.88 (0.47) & 0.91 (0.13) & 0.64 (0.03) & 1.14 &
81.60 (4.5) & \underline{90.28 (3.8)} & 58.81 (7.0) & 81.54 (2.6) & 80.19 (13.1) & 78.48 \\
HiGNN & 7.11 (1.25) & 4.51 (0.94) & 1.68 (0.05) & 4.44 &
\underline{84.54 (2.3)} & 86.12 (1.3) & 60.28 (3.8) & \underline{81.57 (1.8)} & 79.54 (16.7) & 78.41 \\
ResGAT & 1.89 (0.44) & 1.09 (0.16) & 0.71 (0.03) & 1.23 &
75.49 (7.9) & 87.11 (5.0) & \underline{61.40 (2.9)} & 79.83 (3.5) & 81.09 (3.3) & 76.98 \\
ChemBERTa & 2.96 (0.41) & 1.13 (0.19) & 0.80 (0.03) & 1.63 &
81.41 (4.0) & 88.43 (4.7) & 60.80 (1.9) & 78.48 (1.6) & 83.90 (6.2) & 78.60 \\
HiMol & 2.93 (0.28) & 0.87 (0.05) & 0.70 (0.04) & 1.50 &
82.44 (4.2) & 88.86 (4.8) & 57.78 (4.3) & 80.81 (1.7) & 66.19 (5.6) & 75.22 \\
MolCLR & 2.47 (0.53) & 1.28 (0.08) & 0.65 (0.05) & 1.47 &
82.84 (3.4) & 87.66 (4.6) & 58.13 (1.5) & 78.43 (1.6) & 85.74 (3.6) & 78.56 \\
Uni-Mol & 1.73 (0.37) & \underline{0.78 (0.08)} & \textbf{0.57 (0.04)} & \underline{1.03} &
83.57 (3.7) & 87.86 (3.7) & 61.23 (1.6) & 81.40 (1.5) & 80.73 (8.8) & 78.96 \\
TopExpert & \textemdash & \textemdash & \textemdash & \textemdash &
77.20 (4.8) & 87.05 (3.9) & 59.61 (3.5) & 77.68 (2.3) & 75.82 (7.6) & 75.47 \\
DA-MoE & 2.78 (0.71) & 1.13 (0.20) & 0.69 (0.05) & 1.53 &
79.60 (7.7) & 86.83 (5.2) & 55.61 (3.3) & 74.00 (2.2) & 75.11 (8.7) & 74.23 \\
\cmidrule{1-11}
\textbf{Our models} &  &  &  &  &  & &  &  & & \\
MI-MoE-GIN & 2.33 (0.38) & 0.94 (0.15) & 0.70 (0.04) & 1.32 &
\textbf{85.15 (1.8)} & \textbf{90.46 (3.5)} & 60.72 (1.4) & 80.39 (2.3) & 89.04 (8.1) & \underline{81.15} \\
MI-MoE-SchNet & 1.76 (0.37) & \underline{0.78 (0.13)} & 0.63 (0.04) & 1.06 &
82.19 (4.2) & 89.69 (4.9) & \textbf{61.42 (2.4)} & 80.43 (2.0) & \textbf{92.71 (3.9)} & \textbf{81.29} \\
MI-MoE-ViSNet & \underline{1.58 (0.24)} & \textbf{0.72 (0.10)} & 0.62 (0.04) & \textbf{0.97} &
82.37 (4.0) & 88.68 (5.3) & 59.72 (2.2) & 78.14 (3.2) & 89.39 (4.8) & 79.66 \\
MI-MoE-PaiNN & \textbf{1.52 (0.28)} & \underline{0.78 (0.11)} & \underline{0.61 (0.02)} & \textbf{0.97} &
81.31 (5.2) & 87.95 (2.9) & 58.78 (2.3) & 80.69 (2.7) & \underline{90.85 (4.6)} & 79.92 \\
\bottomrule
\end{tabular}
\end{adjustbox}
\label{tab:result_all}
\vspace{-.3cm}
\end{table*}

\subsection{Polymer Property Prediction Tasks}
\subsubsection{Datasets}
We evaluate our method on four polymer property prediction tasks~\cite{kuenneth2021polymer}, including electron affinity $E_{ea}$, ionization energy $E_i$, crystallization tendency $X_c$, and refractive index $\eta_c$. Following prior work~\cite{shen2024molecular}, we represent each polymer using its repeat unit (monomer) and adopt the same dataset splitting protocol. Additional dataset details are provided in the Appendix.

\subsubsection{Baselines}
We compare our approach against six state-of-the-art polymer and molecular property prediction models, following the experimental protocol in~\cite{shen2024molecular}. The evaluated baselines include polyBERT~\cite{kuenneth2023polybert}, TransPolymer~\cite{adams2008engineering}, polyGNN~\cite{gurnani2023polymer}, Mol-GDL~\cite{shen2023molecular}, GEM~\cite{fang2022geometry}, and Mol-TDL~\cite{shen2024molecular}. Additional model details are in the Appendix.

\subsubsection{Performance Evaluation}

\begin{table}[!ht]
\centering
\caption{Average RMSE results (including standard deviations) on the polymer property prediction tasks for $E_{ea}$, $E_{i}$, $X_{c}$, and $\eta_{c}$. Best results are in bold; second-best are underlined.}
\resizebox{0.9\linewidth}{!}{
\begin{tabular}{l|cccc}
\toprule
Model 
& $E_{ea}$ (RMSE $\downarrow$) 
& $E_{i}$ (RMSE $\downarrow$)
& $X_{c}$ (RMSE $\downarrow$)
& $\eta_{c}$ (RMSE $\downarrow$) \\
\midrule
polyBERT
& $0.308 \pm 0.004$ 
& $0.525 \pm 0.011$ 
& $17.646 \pm 0.220$ 
& $0.131 \pm 0.004$ \\

TransPolymer
& $0.320 \pm 0.020$ 
& $0.390 \pm 0.070$ 
& $16.570 \pm 0.680$ 
& $0.100 \pm 0.020$ \\

polyGNN 
& $0.341 \pm 0.055$ 
& $0.540 \pm 0.170$ 
& $18.600 \pm 1.900$ 
& $0.093 \pm 0.030$ \\

Mol-GDL
& $0.552 \pm 0.596$ 
& $0.563 \pm 0.024$ 
& $18.768 \pm 0.858$ 
& $0.086 \pm 0.004$ \\

GEM 
& $0.274 \pm 0.050$ 
& $\underline{0.313 \pm 0.016}$ 
& $17.820 \pm 1.684$ 
& $0.092 \pm 0.023$ \\

Mol-TDL 
& $\underline{0.263 \pm 0.013}$ 
& $0.417 \pm 0.021$ 
& $15.862 \pm 0.825$ 
& $\underline{0.068 \pm 0.006}$ \\

\midrule
MI-MoE-GIN & $0.315 \pm 0.017$ & $0.848 \pm 0.227$ & $10.369 \pm 1.382$ & $0.118 \pm 0.015$ \\
MI-MoE-GTN & $0.313 \pm 0.014$ & $0.465 \pm 0.051$ & $\underline{9.317 \pm 0.136}$ & $0.153 \pm 0.021$ \\
MI-MoE-SchNet & \textbf{0.148 $\pm$ 0.013} & \textbf{0.240 $\pm$ 0.007} & \textbf{8.945 $\pm$ 0.376} & \textbf{0.065 $\pm$ 0.008} \\
MI-MoE-PaiNN & $0.314 \pm 0.010$ & $0.389 \pm 0.027$ & $13.291 \pm 0.560$ & $0.086 \pm 0.001$ \\

\bottomrule
\end{tabular}
}
\label{tab:result_polymer}
\vspace{-.3cm}
\end{table}

Table~\ref{tab:result_polymer} reports the average RMSE performance of all methods on four polymer property prediction tasks. Among the baseline models, Mol-TDL and GEM achieve strong results across multiple properties, with Mol-TDL performing particularly well on $E_{ea}$ and $\eta_c$, and GEM yielding the lowest RMSE on $E_i$ among non–MI-MoE methods. In contrast, sequence-based and graph-based baselines such as polyBERT, TransPolymer, and polyGNN exhibit comparatively higher errors, especially on structure-sensitive properties. Incorporating the proposed MI-MoE framework consistently improves predictive performance across most tasks and backbones. In particular, MI-MoE-SchNet achieves the best overall performance, substantially reducing RMSE on all four properties and outperforming both traditional baselines and strong geometry-aware models. Notably, MI-MoE-SchNet reduces the RMSE on $X_c$ from 15.86 (Mol-TDL) to 8.95 and on $E_i$ from 0.31 (GEM) to 0.24, highlighting the effectiveness of multiscale expert routing. These results demonstrate that MI-MoE enables more accurate polymer property prediction by adaptively modeling molecular interactions across multiple geometric regimes.

\subsection{Model Analysis}
\label{subsec:model_analysis}

\subsubsection{Effect of Interaction Cutoff Selection}
\label{subsec:cutoff_ablation}

\begin{figure}[!ht]
\centering
\begin{subfigure}[b]{0.9\linewidth}
\centering
\includegraphics[width=1\linewidth]{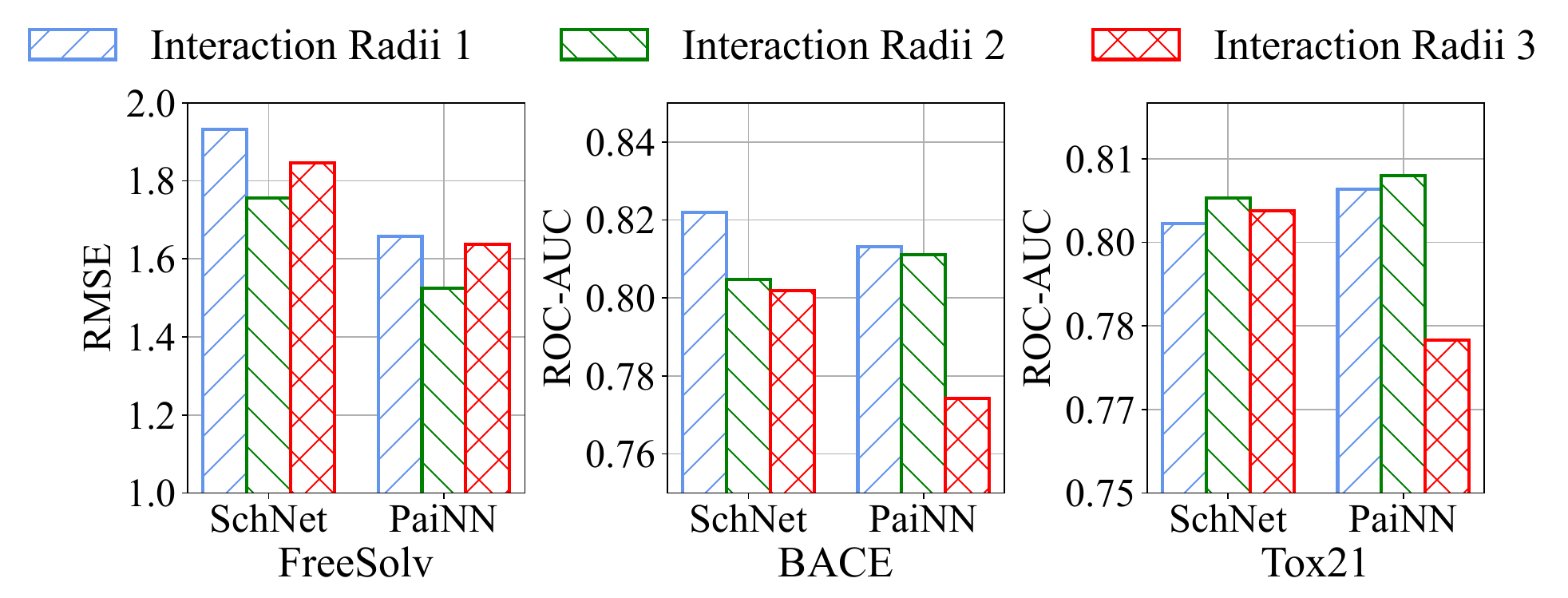}
\end{subfigure}
\caption{MI-MoE performance across interaction radii.}
\label{fig:ablation_cutoff}
\end{figure}

In addition to the default interaction radii
$\{c_k\}_{k=1}^{K} = \{2.0,\, 2.5,\, 3.0,\, 3.5,\, 4.0\}\,\text{\AA}$,
we further evaluate two alternative cutoff configurations,
$\{2.0,\, 3.0,\, 4.0,\, 5.0,\, 6.0\}\,\text{\AA}$ and
$\{2.0,\, 4.0,\, 6.0,\, 8.0,\, 10.0\}\,\text{\AA}$,
to assess the sensitivity of MI-MoE to the choice of interaction scales.
Figure~\ref{fig:ablation_cutoff} reports the performance of MI-MoE-SchNet and
MI-MoE-PaiNN on the FreeSolv, BACE, and Tox21 datasets.
On the BACE dataset, the first cutoff configuration consistently yields the best
performance for both SchNet and PaiNN backbones.
In contrast, FreeSolv and Tox21 achieve their highest performance under the
second cutoff configuration, indicating that the optimal interaction scale
depends on dataset-specific characteristics.

\subsubsection{Effect of Gating Network Architecture}
\label{subsec:gate_arch}

\begin{table}[!ht]
\centering
\caption{MI-MoE performance across topological gating networks.}
\resizebox{0.75\linewidth}{!}{
\begin{tabular}{l|c|ccc}
\toprule
Model & Gating Net & FreeSolv & BACE & Tox21 \\
\midrule
SchNet & - & 1.91 (0.21) & 74.38 (4.9) & 77.59 (2.5) \\
+MI-MoE & MLP & \textbf{1.76 (0.37)} & \textbf{82.19 (4.2)} & \textbf{80.43 (2.0)} \\
+MI-MoE & Transformers & 1.86 (0.48) & 81.71 (4.7) & 80.38 (2.1) \\
\midrule
PaiNN & - & 1.59 (0.41) & 77.69 (4.7) & 78.25 (3.1) \\
+MI-MoE & MLP & \textbf{1.52 (0.28)} & 81.31 (5.2) & \textbf{80.69 (2.7)} \\
+MI-MoE & Transformers & 1.69 (0.42) & \textbf{83.32 (4.1)} & 80.17 (2.1) \\
\bottomrule
\end{tabular}
}
\label{tab:result_gate_net}
\end{table}

Table~\ref{tab:result_gate_net} compares different topological gating network designs within the MI-MoE framework. Both MLP- and Transformer-based gates consistently outperform the corresponding single-scale SchNet and PaiNN baselines across regression and classification tasks, demonstrating the effectiveness of topology-aware routing. For SchNet, the MLP gate achieves the strongest overall performance, while for PaiNN, the Transformer-based gate yields the best BACE results. Overall, these results indicate that simpler MLP gates provide competitive and stable gains, whereas attention-based gates offer additional flexibility for capturing dataset-specific interaction patterns.

\subsubsection{Effect of Expert Configuration}
\label{subsec:expert_config}

\begin{table}[!ht]
\centering
\caption{MI-MoE performance across different expert configurations.}
\label{tab:result_dense_and_1_expert}
\resizebox{0.75\linewidth}{!}{
\begin{tabular}{l|ccc}
\toprule
Model & FreeSolv & BACE & Tox21 \\
\midrule
SchNet & 1.91 (0.21) & 74.38 (4.9) & 77.59 (2.5) \\
+ MI-MoE (One Expert) & 1.77 (0.37) & 81.67 (1.6) & 80.15 (2.2) \\
+ MI-MoE (Dense) & 1.90 (0.52) & 81.35 (4.8) & 77.94 (2.1) \\
+ MI-MoE (Sparse) & \textbf{1.76 (0.37)} & \textbf{82.19 (4.2)} & \textbf{80.43 (2.0)} \\
\midrule
PaiNN & 1.59 (0.41) & 77.69 (4.7) & 78.25 (3.1) \\
+ MI-MoE (One Expert) & 1.74 (0.24) & 62.86 (17.6) & 79.50 (2.6) \\
+ MI-MoE (Dense) & 1.58 (0.30) & 61.36 (15.9) & 77.89 (2.2) \\
+ MI-MoE (Sparse) & \textbf{1.52 (0.28)} & \textbf{81.31 (5.2)} & \textbf{80.69 (2.7)} \\
\bottomrule
\end{tabular}
}
\end{table}

Table~\ref{tab:result_dense_and_1_expert} compares different MI-MoE
configurations, including a single-expert variant, dense expert aggregation, and sparse expert selection. Using only one expert already improves upon the corresponding single-scale
baselines for SchNet, indicating that topology-aware routing itself contributes positively to performance.
However, the full sparse MI-MoE consistently achieves the best results across
all datasets, yielding the lowest RMSE on FreeSolv and the highest ROC-AUC on BACE and Tox21 for both SchNet and PaiNN. In contrast, dense expert aggregation provides limited or unstable gains, particularly for PaiNN, suggesting that indiscriminate expert averaging can dilute scale-specific information.
These results highlight the importance of sparse, topology-guided expert selection in effectively leveraging multiscale interaction modeling.

\section{Conclusion}
\label{sec:conclusion}

In this work, we identify a key limitation of many existing 3D molecular graph neural networks, namely their reliance on a single, globally fixed interaction cutoff, which can be suboptimal for molecules with diverse structural characteristics. To address this limitation, we propose \emph{MI-MoE}, a plug-and-play multiscale mixture of experts framework built upon distance-driven graph filtrations. MI-MoE evaluates multiple 3D GNN experts on cutoff-specific interaction graphs and employs a topology-aware gating mechanism, guided by normalized topological indices and persistent homology-derived Betti curves, to adaptively combine expert predictions. Although MI-MoE operates on 3D molecular graphs, it requires only SMILES strings as input, with conformers generated on the fly using RDKit, enabling practical deployment without sacrificing predictive performance. Experiments on molecular and polymer property prediction benchmarks demonstrate consistent improvements across invariant and equivariant backbones, while ablation studies highlight the importance of topology-guided, sparse expert routing. Future work will explore the integration of higher-precision conformers, as well as more expressive 3D GNN experts, and pretraining strategies to further enhance robustness and generalization.

\bibliographystyle{unsrt}  
\bibliography{references}  


\newpage
\appendix

\section{Expert Model Description}
\label{appen:expert_model}

The proposed MI-MoE framework is designed as a plug-and-play module and can be integrated with a variety of 3D graph neural network architectures, which serve as expert models. In our experiments, we evaluate the following representative backbones:

\begin{itemize}
    \item \textbf{Invariant 3D GNNs:}
    \begin{itemize}
        \item \textbf{SchNet}~\cite{schutt2017schnet}, which models interatomic interactions using continuous-filter convolutions parameterized by pairwise distances, ensuring invariance to rigid transformations.
        \item \textbf{Graph Isomorphism Network (GIN)}~\cite{xu2018how} and \textbf{Graph Transformer Network (GTN)}~\cite{shi2021masked}, adapted to incorporate geometric information while preserving invariance to rotations and translations in the same way as \textbf{SchNet}.
        \item \textbf{DimeNet++}~\cite{gasteiger2020fast}, which explicitly encodes angular information through directional message passing to capture higher-order geometric interactions.
        \item \textbf{ViSNet}~\cite{wang2024enhancing}, a geometry-aware architecture that leverages coupled scalar and vector features to model complex spatial dependencies.
    \end{itemize}

    \item \textbf{Equivariant 3D GNNs:}
    \begin{itemize}
        \item \textbf{EGNN}~\cite{satorras2021n}, which enforces equivariance to Euclidean transformations by jointly updating node features and atomic coordinates during message passing.
        \item \textbf{PaiNN}~\cite{schutt2021equivariant}, which propagates scalar and vector features in an equivariant manner, preserving rotational and translational symmetries throughout the network.
    \end{itemize}
\end{itemize}

To ensure a fair comparison and manageable computational cost within the mixture of experts framework, the depth of each 3D GNN expert is set to be shallower than that of its corresponding standalone baseline, which is described in Table~\ref{tab:depth_settings}. 

\begin{table*}[!ht]
\centering
\caption{Depth settings for standalone 3D GNN baselines and their corresponding MI-MoE expert models. Expert models use reduced depth to ensure comparable computational cost within the mixture of experts framework.}
\label{tab:depth_settings}
\begin{tabular}{lcc}
\toprule
Backbone Model & Standalone Depth & Expert Depth (MI-MoE) \\
\midrule
SchNet        & 6  & 3 \\
GIN           & 4  & 3 \\
GTN           & 4  & 3 \\
DimeNet++     & 4  & 3 \\
ViSNet        & 6  & 3 \\
EGNN          & 7  & 3 \\
PaiNN         & 3  & 3 \\
\bottomrule
\end{tabular}
\end{table*}

\paragraph{Other GNN Architectures.}
We emphasize that the proposed framework is agnostic to the choice of backbone and can be extended to incorporate more advanced 3D GNN architectures, including
GemNet~\cite{klicpera2021gemnet} and MGNN~\cite{chang2025mgnn}.
Due to computational constraints and limited experimental resources, we leave a comprehensive evaluation of these architectures for future work.

\section{Dataset Description}\label{appen:dataset}

\subsection{Molecular Property Prediction Tasks}

\subsubsection{Benchmark Datasets}

{To evaluate the effectiveness of our proposed framework, we performed extensive experiments on eight benchmark datasets for molecular properties prediction, which include  FreeSolv, ESOL, Lipophilicity, BACE, BBBP, ClinTox, SIDER, and Tox21. These datasets were obtained from MoleculeNet \cite{wu2018moleculenet}, a well-established resource for benchmarking ML approaches in this domain. After data acquisition, we carried out a thorough curation process to exclude unqualified samples, resulting in a reduced number of entries in the final datasets \cite{nguyen2024resgat}.}

{For all the MoleculeNet datasets, we employed the scaffold splitting method. Scaffold splitting, which separates molecules based on their molecular substructures, i.e., scaffolds, offers a more rigorous evaluation of a model's ability to generalize to out-of-distribution samples compared to the normal random splitting. Molecules were split into training and testing sets in a 9:1 ratio, with the training set further divided into training and validation sets (also in a 9:1 ratio) for hyperparameter tuning. In line with the MoleculeNet benchmark, we used the root mean squared error (RMSE) metric to evaluate our approach for regression datasets, which are ESOL, FreeSolv, and Lipophilicity. For classification datasets including BACE, BBBP, ClinTox, SIDER, and Tox21, the metric is the area under the receiver operating characteristic curve (ROC-AUC). To ensure robust evaluation, we conducted five trials for each splitting method with different random seeds, reporting the average performance and standard deviation to assess both consistency and variability across data splits.}
\begin{table*}[t]
\centering
\caption{Details of the MoleculeNet datasets.}
\label{tab:moleculenet}
\begin{tabular}{lclccl}
\toprule
{Dataset} & {$\#$tasks} & {Task type} &  {$\#$original samples} & {$\#$refined samples}  \\ 
\midrule
 ESOL     & 1 & & 1128 & 1115  \\ 
 FreeSolv  & 1&  Regression  & 642  & 635   \\
 Lipophilicity      & 1 &  & 4200 & 4100  \\ 
\midrule
 BACE    & 1 & Binary classification & 1513 & 1454  \\ 
 BBBP    & 1 &  & 2050 & 1760  \\ 
 \midrule
ClinTox & 2 & & 1484  & 1349   \\ 
SIDER   & 27 & Multi-label classification & 1427 & 1225  \\ 
Tox21   & 12 & & 7831 & 7381  \\ 
\bottomrule
\end{tabular}
\end{table*}

\subsubsection{Data Curation}

{Before conducting experiments, we had carefully curated the molecular data to ensure it met the necessary quality standards for model development and evaluation. This process primarily focused on the MoleculeNet datasets~\cite{wu2018moleculenet}. The curation workflow, comprised four key stages: $(i)$ Validation, $(ii)$ Cleaning, $(iii)$ Normalization, and $(iv)$ Final Verification. {First, all SMILES data were standardized into their canonical forms.}  
During the Validation stage, we excluded molecules categorized as inorganics, mixtures, or organometallics. During the Cleaning stage, salts and charged molecules were filtered out, with metal-containing charged molecules being excluded and organic charged molecules undergoing neutralization. Although neutralizing organic charged molecules is a subject of ongoing debate due to difficulties in evaluating their experimental activity, this step was implemented to ensure data consistency.  
The Normalization phase involved detautomerization, destereoisomerization, and chemotype unification, standardizing tautomers, stereoisomers, and chemotypes into their canonical forms and eliminating duplicates. Finally, in the Final Verification stage, molecules with conflicting labels were resolved manually: structures with inconsistent labels or CAS numbers were excluded, while duplicates with consistent labels were merged. PubChem~\cite{Kim2020PubChemI2} and ChEMBL~\cite{Zdrazil2023ChEMBL} databases were used for the final structural validation. Table~\ref{tab:moleculenet} summarizes the details of the curated MoleculeNet datasets.}

\subsection{Polymer Property Prediction Tasks}

\begin{table*}[!ht]
\centering
\caption{Details of the datasets for polymer property prediction.}
\label{tab:polymer_datasets}
\begin{tabular}{l l c c c c}
\toprule
Dataset & Property & Task type & $\#$ samples \\
\midrule
$E_{ea}$ & Electron affinity & Regression & 368 \\
$E_{i}$  & Ionization energy & Regression & 370 \\
$X_{c}$  & Crystallization tendency & Regression & 432 \\
$n_{c}$ & Refractive index & Regression & 382 \\
\bottomrule
\end{tabular}
\end{table*}

\subsubsection{Benchmark Datasets}
To evaluate the effectiveness of our proposed framework, we evaluated our method on four polymer property prediction tasks~\cite{kuenneth2021polymer}, including electron affinity $E_{ea}$, ionization energy $E_i$, crystallization tendency $X_c$, and refractive index $\eta_c$. Following prior work~\cite{shen2024molecular}, we represent each polymer using its repeat unit, monomers, to construct 3D molecular representations for property prediction. All datasets are randomly split into training, validation, and test sets with an 8:1:1 ratio.

\section{Baseline Model Description}
\label{appen:baseline_model}
\subsection{Molecular Property Prediction Tasks}

We compare our approach against a diverse set of representative models for molecular property prediction and graph-level learning, spanning SMILES-based language models, graph neural networks, mixture of experts architectures, hierarchical representations, and 3D geometric learning frameworks.
\begin{itemize}

    \item \textbf{AttentiveFP}~\cite{xiong2019pushing} is an attention-based GNN that applies graph attention mechanisms at both atom and molecule levels. By allowing nonlocal interactions to influence representation learning, AttentiveFP captures long-range intramolecular effects such as hydrogen bonding, while also offering interpretability through attention-weight visualization over chemically relevant substructures.

    \item \textbf{D-MPNN}~\cite{yang2019analyzing} is a directed message-passing neural network that operates on directed bonds rather than atoms, mitigating redundant message loops (totters) that can introduce noise in standard MPNNs. It is often augmented with fixed molecular descriptors, which enhances robustness, particularly in low-data regimes.

    \item \textbf{HiGNN}~\cite{zhu2022hignn} employs hierarchical co-representation learning by jointly modeling atomic graphs and chemically meaningful BRICS fragments. Its feature-wise attention mechanism adaptively recalibrates atomic feature channels, enabling the model to highlight structurally and functionally important molecular components for property prediction.

    \item \textbf{ResGAT}~\cite{nguyen2024resgat} enhances graph attention networks with residual connections to stabilize optimization and increase representational capacity. By incorporating both block-level and layer-integrated residuals, ResGAT achieves improved training efficiency and competitive performance across classification and regression tasks.

    \item \textbf{HiMol}~\cite{zang2023hierarchical} is a self-supervised framework based on a Hierarchical Molecular GNN that explicitly models atom-, motif-, and graph-level representations. Molecules are decomposed into motifs and augmented with a global node to facilitate bidirectional information flow. Multi-level pretraining tasks encourage the learning of rich chemical semantics across different structural scales.

    \item \textbf{MolCLR}~\cite{wang2022molecular} is a contrastive self-supervised learning framework that learns molecular representations from large unlabeled datasets via graph augmentations, including atom masking, bond deletion, and subgraph removal. By maximizing agreement between augmented views of the same molecule, MolCLR enables simple GNN backbones to learn robust and chemically meaningful features.

    \item \textbf{ChemBERTa}~\cite{chithrananda2020chemberta} is a transformer-based molecular representation model adapted from RoBERTa and pretrained using masked language modeling on large-scale SMILES corpora. By leveraging massive chemical databases such as PubChem, ChemBERTa learns contextualized token representations that capture statistical regularities of chemical syntax and semantics, enabling effective transfer to downstream molecular classification and regression tasks.

    \item \textbf{Uni-Mol}~\cite{zhou2023unimol} is a large-scale 3D molecular representation learning framework based on SE(3)-equivariant Transformers. Pretrained on hundreds of millions of molecular conformations and protein pockets, Uni-Mol directly processes atomic coordinates and is particularly effective for tasks requiring explicit geometric reasoning, such as protein–ligand binding and conformation generation.

    \item \textbf{TopExpert}~\cite{kim2023learning} introduces topology-aware expert specialization by clustering molecules according to structural and scaffold-level semantics. A gating module assigns each molecule to a topology-specific expert, enabling the model to capture discriminative features tailored to distinct molecular topologies and improving generalization to unseen scaffolds.

    \item \textbf{DA-MoE}~\cite{yao2025moe} addresses the depth sensitivity of graph neural networks by treating GNNs with different numbers of layers as independent experts. A structure-aware gating network selects appropriate experts for each input graph, allowing the model to adaptively match network depth to graph scale and complexity, thereby improving both efficiency and predictive performance.

\end{itemize}

\subsection{Polymer Property Prediction Tasks}

We evaluate our method against a set of representative polymer property prediction models that span topological deep learning, graph neural networks, chemical language models, and geometry-aware architectures.

\begin{itemize}
    \item \textbf{polyBERT}~\cite{kuenneth2023polybert} is a chemical language model that formulates polymer structures as a sequence modeling problem. Built upon a DeBERTa-based Transformer architecture and pretrained on approximately 100 million hypothetical PSMILES strings, polyBERT learns rich contextual embeddings of polymer chemistry. This representation enables highly efficient feature extraction and property prediction, achieving orders-of-magnitude speedups over handcrafted descriptors while maintaining competitive accuracy.

    \item \textbf{TransPolymer}~\cite{adams2008engineering} adopts a Transformer-based language modeling approach for polymer property prediction using SMILES representations. Through large-scale pretraining on millions of augmented polymer sequences, TransPolymer captures complex sequential patterns that implicitly encode spatial and topological information, enabling competitive performance relative to graph-based molecular models.

    \item \textbf{polyGNN}~\cite{gurnani2023polymer} is a multitask graph neural network designed for polymer systems, where polymers are modeled as periodic graphs derived from their repeat units. This formulation ensures invariance to transformations such as translation and unit repetition. By jointly learning multiple correlated polymer properties, polyGNN improves predictive performance in low-data regimes and significantly accelerates feature computation compared to traditional fingerprint-based methods.

    \item \textbf{Mol-GDL}~\cite{shen2023molecular} models molecular topology through a hierarchy of graphs that treat covalent and non-covalent interactions on an equal footing. It employs geometry-derived node features determined solely by atomic types and pairwise Euclidean distances, allowing the model to implicitly learn physical and chemical properties via multiscale structural representations.

    \item \textbf{GEM}~\cite{fang2022geometry} is a geometry-aware GNN that simultaneously models atoms, bonds, and bond angles using a dual-graph architecture. GEM incorporates geometry-level self-supervised pretraining objectives, such as bond length prediction and atomic distance matrix reconstruction, to encode critical three-dimensional spatial information, which is particularly important for accurate polymer property prediction.
    
    \item \textbf{Mol-TDL}~\cite{shen2024molecular} represents polymer repeat units as a family of Vietoris–Rips complexes constructed over multiple interaction thresholds, enabling the joint modeling of covalent and non-covalent interactions within a unified topological framework. Unlike conventional graph-based approaches, Mol-TDL performs message passing on higher-order simplices, such as triangles (2-simplices), using upper and lower adjacency operators to explicitly encode many-body interactions and multiscale topological dependencies.
\end{itemize}

\end{document}